\documentclass[letterpaper]{article} 
\usepackage{aaai2026}  
\usepackage{times}  
\usepackage{helvet}  
\usepackage{courier}  
\usepackage[hyphens]{url}  
\usepackage{graphicx} 
\usepackage{natbib}  
\usepackage{caption} 
\usepackage{algorithm}
\usepackage{algorithmic}

\usepackage{amsmath}
\usepackage{amssymb}
\usepackage{booktabs}
\usepackage{multirow}

\usepackage{newfloat}
\usepackage{listings}
\DeclareCaptionStyle{ruled}{labelfont=normalfont,labelsep=colon,strut=off} 
\floatstyle{ruled}
\newfloat{listing}{tb}{lst}{}
\floatname{listing}{Listing}
\title{Sat2Flow: A Structure-Aware Diffusion Framework for Human Flow Generation from Satellite Imagery}
\author{
    Xiangxu Wang\textsuperscript{\rm 1},
    Tianhong Zhao\textsuperscript{\rm 1},
    Wei Tu\textsuperscript{\rm 2},
    Bowen Zhang\textsuperscript{\rm 1},
    Guanzhou Chen\textsuperscript{\rm 3},
    Jinzhou Cao\textsuperscript{\rm 1}\thanks{Corresponding author.}\\
}
\affiliations{
    \textsuperscript{\rm 1}College of Big Data and Internet Shenzhen Technology University, China\\
    \textsuperscript{\rm 2}Department of Urban Informatics, School of Architecture and Urban Planning, Shenzhen University, China\\
    \textsuperscript{\rm 3}State Key Laboratory of Information Engineering in Surveying, Mapping, and Remote Sensing, Wuhan University, China \\

    wangxiangxu2023@email.szu.edu.cn, \{caojinzhou, zhaotianhong\}@sztu.edu.cn,\\
    tuwei@szu.edu.cn, zhang\_bo\_wen@foxmail.com, cgz@whu.edu.cn
}

\usepackage{bibentry}

\begin{document}

\maketitle

\begin{abstract}
Origin-Destination (OD) flow matrices are critical for urban mobility analysis, supporting traffic forecasting, infrastructure planning, and policy design. Existing methods face two key limitations: (1) reliance on costly auxiliary features (e.g., Points of Interest, socioeconomic statistics) with limited spatial coverage, and (2) fragility to spatial topology changes, where reordering urban regions disrupts the structural coherence of generated flows. 
We propose Sat2Flow, a structure-aware diffusion framework that generates structurally coherent OD flows using only satellite imagery. Our approach employs a multi-kernel encoder to capture diverse regional interactions and a permutation-aware diffusion process that maintains consistency across regional orderings. Through joint contrastive training linking satellite features with OD patterns and equivariant diffusion training enforcing structural invariance, Sat2Flow ensures topological robustness under arbitrary regional reindexing. Experiments on real-world datasets show that Sat2Flow outperforms physics-based and data-driven baselines in accuracy while preserving flow distributions and spatial structures under index permutations.
Sat2Flow offers a globally scalable solution for OD flow generation in data-scarce environments, eliminating region-specific auxiliary data dependencies while maintaining structural robustness for reliable mobility modeling.
\end{abstract}

\begin{links}
    \link{Code}{https://github.com/ai4city-sztu/Sat2Flow}
\end{links}

\section{Introduction}

Accurate generation of origin-destination (OD) flows is fundamental to urban mobility planning~\cite{joubert2020activity}, congestion management~\cite{iqbal2014development}, and public resource allocation~\cite{rong2023complexity}. 
Traditional OD modeling relies heavily on infrastructure sensors (e.g., traffic counters) and travel surveys, which suffer from sparse spatial coverage and high deployment costs~\cite{calabrese2011estimating}.
Early physical models, such as gravity models~\cite{barbosa2018human} operating at the pair-wise level, provided theoretical frameworks but oversimplified urban complexity. With the advancement of machine learning, data-driven approaches have significantly improved prediction accuracy by leveraging matrix-wise level features as inputs, yet they face critical bottlenecks in data accessibility and structural modeling constraints~\cite{rong2023complexity, yan2024urbanclip}.
Current state-of-the-art frameworks exhibit two fundamental limitations:

\textbf{Feature Dependency Dilemma.}
Existing models universally require socio-economic indicators (e.g., population density) and points-of-interest (POI) distributions as essential inputs~\cite{cao2025disentangling, cao2025semigps}. These auxiliary data sources are notoriously difficult to acquire at scale due to privacy restrictions, collection costs, and inconsistent availability across different geographic regions, particularly in developing countries where urban mobility modeling is most urgently needed. In contrast, satellite imagery provides globally accessible remote sensing data that captures essential urban spatial patterns~\cite{mohanty2020deep} without requiring region-specific collection or privacy-sensitive information, offering a scalable alternative to feature-dependent approaches.

\begin{figure}[t]
\centering
\includegraphics[width=0.87\columnwidth]{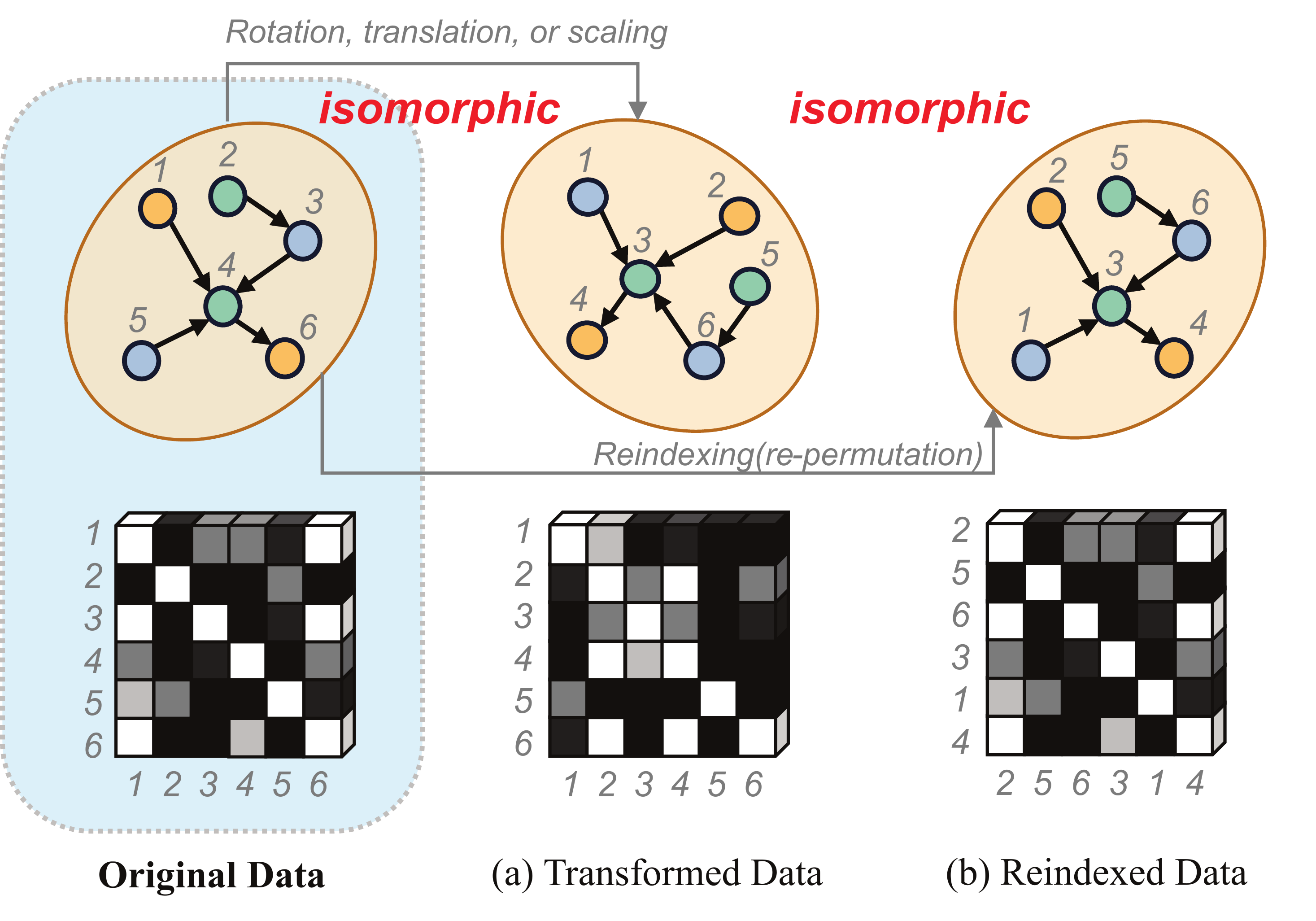}
\caption{Structural consistency in urban OD flows. Geometric transformations (a) and index permutations (b) preserve isomorphic spatial relationships, maintaining structural consistency under regional reindexing.}
\label{fig:augment}
\end{figure}

\textbf{Structural Consistency Deficiency.}
As illustrated in Figure~\ref{fig:augment}, urban OD flow matrices should exhibit structural consistency under different data transformations. When the same urban area undergoes geometric transformations such as rotation, translation, or scaling (Figure~\ref{fig:augment}a), the spatial relationships remain isomorphic to the original configuration.
In computational representation, this geometric invariance corresponds to index permutation—reassigning the regional index while maintaining identical connectivity patterns (Figure~\ref{fig:augment}b).
However, conventional models fail to recognize this equivalence, treating permuted representations as distinct inputs despite encoding identical mobility relationships. This fundamental limitation reveals the absence of structural consistency (i.e., permutation equivariance in computational representation) in existing approaches. 

To address these fundamental limitations, we propose \textbf{Sat2Flow} – a permutation-equivariant framework for generating urban OD flows using globally accessible satellite imagery as the sole input modality. This approach eliminates dependency on scarce auxiliary data by extracting multi-scale urban patterns directly from remote sensing imagery, while ensuring structural consistency and enabling reliable deployment across diverse urban environments worldwide.
The framework operates through three sequential stages: The first stage establishes a systematic pipeline for downloading satellite imagery of urban regions and encoding them into feature representations. The second stage processes the encoded region features through multi-kernel computing to generate a multi-kernel (MK) matrix, then employs two architecturally identical encoders to align the MK matrix and OD matrix in a shared latent space through pretraining the proposed \textbf{Modality-Joint Contrastive Module}. The third stage utilizes the aligned regional multi-kernel representation from the latent space, along with permutation embedding, as conditional inputs to a novel \textbf{Conditional Latent Diffusion Module}. This module progressively denoises latent variables to synthesize structurally consistent origin-destination flow matrices.
Our contributions are:
\begin{itemize}
\item We present the first OD flow generation solution using only satellite imagery, eliminating dependencies on scarce region-specific auxiliary data while maintaining competitive performance.
\item We develop an adaptive encoder based on multi-kernel functions for regional representation from satellite imagery, capturing local urban features and global spatial dependencies without manual feature engineering.
\item We introduce a conditional latent diffusion module with permutation embeddings that ensures structural consistency under arbitrary regional index reordering, addressing a key limitation in existing approaches.
\item Experiments on 3,333 U.S urban areas demonstrate the state-of-the-art performance of Sat2Flow, showing superior scalability and global applicability.
\end{itemize}

\section{Primaries}

\begin{figure*}[t]
\centering
\includegraphics[width=0.87\textwidth]{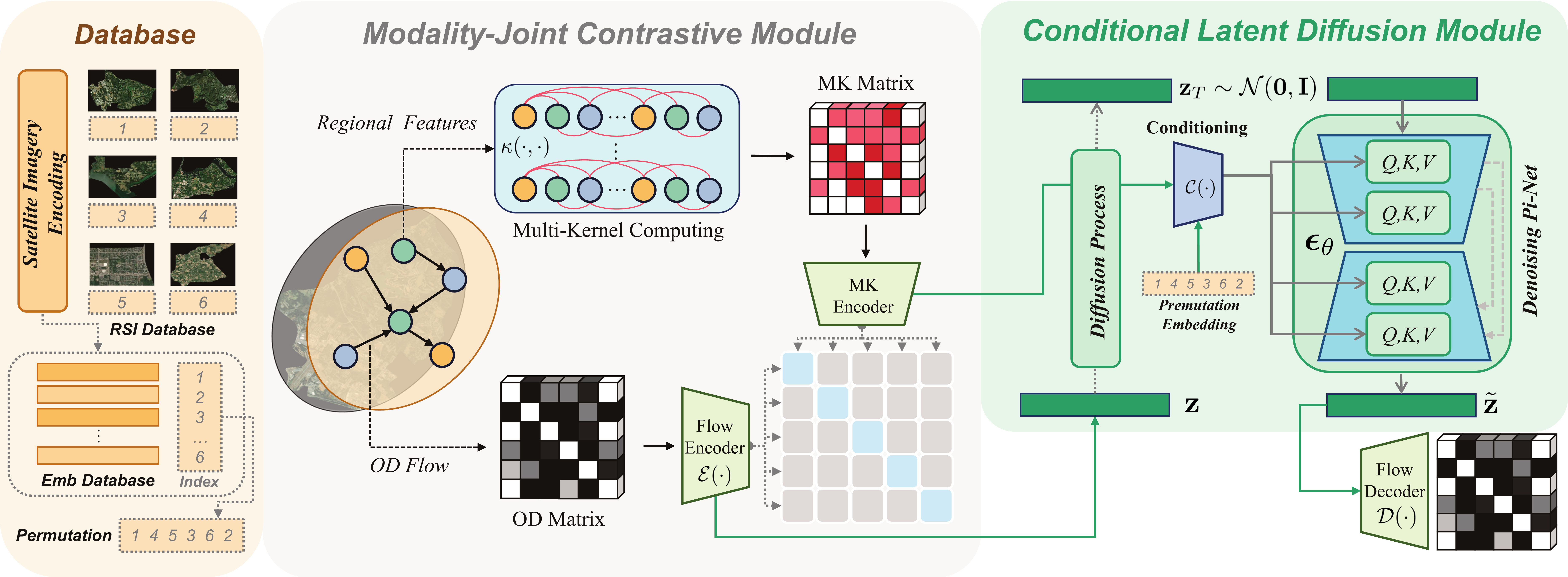}
\caption{Overview of the Sat2Flow framework, featuring three sequential stages (satellite encoding, multi-kernel contrastive learning, and diffusion generation), with the latter two stages integrated within the Modality-Joint Contrastive Module and Conditional Latent Diffusion Module, respectively.}
\label{fig:framework}
\end{figure*}

\noindent\textbf{Definition 1. (Urban Region).}
\textit{A city is partitioned into $N$ distinct urban regions $\mathcal{R} = \{r_i\}_{i=1}^N$, where each region $r_i$ corresponds to a geographically defined unit such as a census tract or administrative division.}

\noindent\textbf{Definition 2. (Satellite Imagery).}
\textit{Satellite imagery $\mathcal{I}$ comprises high-resolution remote sensing data that capture geospatial features across urban landscapes. 
For each region $r_i$,  we extract a georeferenced image $\mathbf{I}_i\in \mathbb{R}^{h\times w\times c}$ with spatial dimensions $h$ and $w$, which preserves multispectral attributes and ground-level structural patterns.}

\noindent\textbf{Definition 2. (OD Flow).}
\textit{Origin-Destination (OD) flow is formalized as a set of ordered triples $\{(r_o, r_d, M_{r_o\to r_d})\}$, where $r_o$ and $r_d$ denote origin and destination regions, and $M_{r_o\to r_d}$ quantifies the intensity of mobility between these regions. This flow can also be represented as a matrix $\mathbf{M}\in\mathbb R^{N\times N}$, or interpreted as a single-channel image with $N\times N$ spatial resolution.}

\noindent\textbf{Definition 3. (Kernel Function).}
\textit{A kernel function $\kappa(\cdot,\cdot)$ is defined on a Reproducing Kernel Hilbert Space (RKHS) $\mathcal{H}$, where it serves as a similarity measure between feature vectors $\mathbf{x}_i, \mathbf{x}_j\in \mathcal{X}$. The kernel satisfies:}
\begin{equation*}
    \kappa(\mathbf x_i, \mathbf x_j)=\langle\phi(\mathbf x_i), \phi(\mathbf x_j)\rangle_{\mathcal H}=\phi(\mathbf x_i)^\top\phi(\mathbf x_j),
\end{equation*}
\textit{where $\phi(\cdot):\mathcal{X}\mapsto\mathcal{H}$ is an implicit feature mapping that embeds the input feature space into RKHS. This formulation ensures that $\kappa(\cdot,\cdot)$ is symmetric and positive semi-definite, aligning with Mercer's theorem and enabling kernel methods for spatial analysis.}

\noindent\textbf{Definition 4. (Permutation Equivariant).}
\textit{A generative model $\mathcal{G}$ is permutation-equivariant if, for any permutation $\pi$ of urban regions, the generated OD matrix $\mathbf{M}'=\mathcal{G}(\{\mathbf x_{\pi(i)}\}_{i=1}^N)$ satisfies:}
\begin{equation*}
    [\mathbf{M}']_{i,j}=[\mathbf{M}]_{\pi(i),\pi(j)},
\end{equation*}
\textit{where $\mathbf{M}=\mathcal{G}(\{\mathbf x_{i}\}_{i=1}^N)$ is the original output matrix. This property ensures that the model's predictions are structurally consistent with input permutations, preserving topological relationships under reordering.}

\noindent\textbf{Problem Formulation (OD Flow Generation).}
\textit{Given a set of regional feature vectors $\{\mathbf{x}_i| r_i\in\mathcal{R}\}_{i=1}^N\in\mathbb R^{N\times d}$, the goal is to learn a permutation-equivariant conditional generative model $\mathcal G$ that reconstructs the OD flow matrix $\mathbf{M}\in\mathbb{R}^{N\times N}$.}

\section{Methodology}
This section elaborates on Sat2Flow, which is a three-stage framework for OD flow generation, as illustrated in Figure~\ref{fig:framework}. 
First, satellite image tiles are collected for urban regions and encoded into feature vectors via a pre-trained vision encoder, forming an embedding database. 
Next, a modality-Joint contrastive module aligns regional multi-kernel representations and OD flow representations in a shared latent space through dedicated encoders and contrastive learning. Finally, a conditional latent diffusion module is trained to generate OD matrices conditioned on the latent presentations and permutation embeddings, ensuring structural consistency under input reordering.

\subsection{Satellite Imagery Encoding}

Sat2Flow constructs a satellite imagery database by systematically collecting and pre-processing georeferenced image tiles for each urban region $r_i\in\mathcal{R}$. The process begins with identifying and downloading high-resolution tiles from standardized imagery services (e.g., Esri World Imagery or Google Earth Engine) that correspond to the geographical boundaries of region $r_i$. To ensure spatial fidelity, these tiles are cropped and aligned with the exact contours of each region while preserving their original regional indices. The indexing scheme establishes a bijective mapping between each image $\mathbf{I}_i\in\mathbb{R}^{h\times w\times c}$ and its associated region $r_i$, ensuring structural consistency during arbitrary re-indexing operations.

The preprocessed satellite imagery is subsequently encoded into semantic feature representations using a pretrained vision encoder. Given the complexity and high dimensionality of raw pixel data, which often contains redundant textures and environmental noise, we adopt RemoteCLIP~\cite{liu2024remoteclip}, a vision-language model specifically designed for satellite imagery analysis. The encoder transforms each image patch $\mathbf{I}_i$ to a high-dimensional feature vector $\mathbf{x}_i$. 
The resulting embeddings $\{\mathbf x_{i}\}_{i=1}^N$ are stored in a structured embedding database with indices preserved from the original satellite image database. These embeddings serve as conditional input for the subsequent multi-kernel encoder and latent diffusion components, where both the regional characteristic vectors $\mathbf{x}_i$ and permutation embeddings are jointly leveraged to generate structurally consistent and topologically robust OD flow matrices.

\subsection{Modality-Joint Contrastive Module}

The second stage of Sat2Flow focuses on cross-modal representation learning through a contrastive pretraining module that jointly encodes OD flows and regional feature vectors. By aligning the latent representations of OD patterns and urban semantics in a shared embedding space, 
serving as the foundation for the subsequent diffusion-based generative modeling.

\subsubsection{Multi-Kernel Encoder}

The complex interactions among urban regions collectively form a city-wide relational network. To capture these latent associations, we propose a \textbf{multi-kernel computing} method that learns region-to-region dependencies through adaptive RKHS mappings~\cite{shawe2004kernel}, distinct from graph neural networks that rely on pre-defined adjacency matrices. 
Specifically, given a set of $L$ mappings $\{\phi_l\}_{l=1}^L$, $L$ kernel functions compute the similarity between two regions $r_i,r_j\in\mathcal R$: 
\begin{equation}
    [\mathbf K]_{i,j}^{l}=\kappa^l(\mathbf{x}_i,\mathbf{x}_j)=\phi_l(\mathbf{x}_i)^\top\phi_l(\mathbf{x}_j),\qquad l=1,2,...,L
\end{equation}
where $\mathbf{x}_i, \mathbf{x}_j\in\mathcal{X}\subseteq\mathbb{R}^d$ are regional feature vectors of $r_i$ and $r_j$, and $\mathbf{K}^l\in\mathbb{R}^{N\times N}$ is represents the $l$-th relational matrix. This yields $L$ kernel matrices $\mathbf{K}^1,\mathbf{K}^2,...,\mathbf{K}^L$, which encode multi-perspective interactions. 

To incorporate structural priors (e.g., road network topology), an additional kernel matrix $\mathbf{K}^{L+1}$ is constructed by leveraging domain knowledge, leading to a tensor $\mathcal{K}\in\mathbb{R}^{N\times N\times (L+1)}$ through concatenation:
\begin{equation}
    \mathcal{K}=\operatorname{Concat}(\mathbf{K}^1,\mathbf{K}^2,...,\mathbf{K}^L, \mathbf{K}^{L+1}).
\end{equation}
This tensor is treated as a multi-channel image with spatial dimensions $N\times N$, where each channel corresponds to a distinct relational prior. Subsequently, an attention-augmented encoder (MK Encoder) is used to process the multichannel relational tensor $\mathcal{K}$, producing a regional multi-kernel representation $\mathbf{z}_c$. This representation serves a dual role: It is first aligned with the OD representation in a shared latent space through contrastive learning, enforcing semantic consistency between regional interaction patterns and mobility dynamics; subsequently, $\mathbf{z}_c$ is integrated as a conditional input to the diffusion module, guiding the generative process by encoding high-level spatial priors.

\subsubsection{Flow Encoder}

The OD flow matrix is interpreted as a single-channel spatial map with resolution $N\times N$, where $N$ varies between cities due to different regional divisions. 
This spatial regularity enables convolutional neural networks to effectively capture hierarchical dependencies. The translation-equivariant property of convolutions further ensures robustness to permutations of region indexing.
We employ an attention-augmented convolutional Variational Autoencoder with a fully convolutional architecture, enabling processing of variable-sized inputs (e.g., cities with differing $N$) through dynamic padding or interpolation. 
An encoder-decoder pair will be trained synchronously, where the encoder $\mathcal{E}(\cdot)$, termed the Flow Encoder, shares the same architecture as the MK Encoder to model the approximate posterior distribution $q_z(\mathbf{z}|\mathbf{M})$ of the latent variable $\mathbf{z}$, while the Flow Decoder $\mathcal{D}(\cdot)$ adopts a symmetric design effectively reconstructing the OD matrix through the generative distribution $p_\theta(\mathbf{M}\vert\mathbf{z})$. 
The encoder and decoder architecture closely resembles the subsequent Pi-net structure detailed in Figure~\ref{fig:pinet}.


\subsubsection{Latent Aligned Training}

To bridge the modality gap between regional features and OD flows, we adopt a CLIP-style modality alignment objective that enforces mutual alignment between latent representations $\mathbf{z}_c$ (from the MK Encoder) and $\mathbf{z}$ (from the Flow Encoder) in the shared latent space. This contrastive loss $\mathcal{L}_{\text{con}}$ encourages positive pairs (matching MK-OD samples) to cluster closely while pushing negative pairs apart, formalized as:
\begin{equation}
\begin{split}
\mathcal{L}_{\text{con}}&= -\frac{1}{2N_B} \left( \sum_{i=1}^{N_B} \log \frac{
\exp\left( \text{sim}(\mathbf{z}^{(i)}, \mathbf{z}_c^{(i)})/\tau \right)}{
\sum_{j=1}^{N} \exp\left( \text{sim}(\mathbf{z}^{(i)}, \mathbf{z}_c^{(j)})/\tau \right)} \right. \\
&\qquad + \left. \sum_{i=1}^{N_B} \log \frac{
\exp\left( \text{sim}(\mathbf{z}_c^{(i)}, \mathbf{z}^{(i)})/\tau \right)}{
\sum_{j=1}^{N_B} \exp\left( \text{sim}(\mathbf{z}_c^{(i)}, \mathbf{z}^{(j)})/\tau \right)} \right),
\end{split}
\end{equation}
where $\text{sim}(\cdot, \cdot)$ denotes the cosine similarity, $\tau$ is the temperature parameter, and $N_B$ represents the number of MK-OD pairs in a mini-batch. 

In parallel, a reconstruction loss is introduced to ensure fidelity in the generative process:
\begin{equation}
\mathcal{L}_{\text{rec}}=- \mathbb{E}_{\mathbf{z} \sim q_z(\mathbf{z}\vert\mathbf{M})} \log p_\theta(\mathbf{M}\vert\mathbf{z}),
\end{equation}
which reduces to a pixel-wise $\ell_2$ loss when the decoder follows an isotropic Gaussian distribution.

To further regularize the latent space, we incorporate a Kullback-Leibler (KL) divergence term that aligns the encoder output distribution $q_z(\mathbf{z}|\mathbf{M})$ with a standard Gaussian prior: 
\begin{equation}
\mathcal{L}_{\text{KL}}=D_{KL}(q_z(\mathbf{z}|\mathbf{M})\|\mathcal{N}(0,\mathbf{I})).
\end{equation}
This encourages the latent representations to remain compact and well-distributed, preventing degenerate solutions during diffusion modeling.

The final training objective combines the contrastive loss $\mathcal{L}_{\text{con}}$, a reconstruction loss $\mathcal{L}_{\text{rec}}$ and the KL loss $\mathcal{L}_{\text{KL}}$:
\begin{equation}
\mathcal{L}=\mathcal{L}_{\text{rec}}+\alpha\mathcal{L}_{\text{con}}+\beta\mathcal{L}_{\text{KL}}
\end{equation}
where $\alpha$ and $\beta$ are hyperparameters that balance the contributions of the contrastive and KL regularization terms relative to the reconstruction loss. 

\subsection{Conditional Latent Diffusion Module}
Building upon the latent representation learned in the contrastive module, this stage introduces a latent diffusion model~\cite{rombach2022high} that generates OD flow matrices conditioned on urban regional multi-kernel representation and permutation embeddings. 

\subsubsection{Permutation Embedding}
To ensure structural coherence in OD generation under arbitrary region permutations, we introduce permutation embeddings that encode explicit structural priors into the denoising process. Formally, we maintain a learnable lookup table $\mathbf{E} \in \mathbb{R}^{N\times d}$, where each row $\mathbf{e}_i$ corresponds to the embedding of the original index $i$. For a given permutation $\pi$, the reordered index sequence $\pi(1), \pi(2), \dots, \pi(N)$ retrieves the corresponding embeddings:  
$$
\mathbf{z}_p = \text{MLP}\left(\text{Concat}\left(\mathbf{e}_{\pi(1)}, \dots, \mathbf{e}_{\pi(N)}\right)\right),
$$  
where $\mathbf{z}_p\in \mathbb{R}^{d_c}$ serves as the permutation-aware condition for the diffusion model. This design explicitly guides the generator to respect the structural relationships among regions during OD flow generation.

\subsubsection{Denoising Pi-Net Backbone}

The diffusion process follows a standard Markov chain formulation, where noise is progressively added over $T$ steps to transform the latent OD representation into a sample from the prior distribution $\mathbf{z}_T\sim\mathcal{N}(0,\mathbf{I})$:
\begin{equation}
    q(\mathbf{z}_t \vert \mathbf{z}_{t-1}) = \mathcal{N}(\mathbf{z}_t; \sqrt{1 - \beta_t} \mathbf{z}_{t-1}, \beta_t\mathbf{I}),\quad t=1,2,...,T.
\end{equation}
The reverse process, governed by the neural backbone, aims to recover $\tilde{\mathbf{z}}=\mathbf{z}$ from $\mathbf{z}_T$ and conditions $\mathcal{C}=\{\mathbf{z}_c, \mathbf{z}_p\}$ by learning to estimate the noise added at each step:
\begin{equation}
    p_\theta(\mathbf{z}_{t-1} \vert \mathbf{z}_t,\mathcal{C}) = \mathcal{N}(\mathbf{z}_{t-1}; \boldsymbol{\mu}_\theta(\mathbf{z}_t, \mathcal{C}, t), \boldsymbol{\Sigma}_\theta(\mathbf{z}_t, \mathcal{C}, t)),
\end{equation}
where 
\begin{equation}
    \boldsymbol{\mu}_\theta(\mathbf{z}_t, \mathcal{C}, t)={\frac{1}{\sqrt{\alpha_t}} \Big( \mathbf{z}_t - \frac{1 - \alpha_t}{\sqrt{1 - \bar{\alpha}_t}} \boldsymbol{\epsilon}_\theta(\mathbf{z}_t, \mathcal{C}, t) \Big)},
\end{equation}
with $\alpha_t=1-\beta_t$, and $\bar{\alpha}_t=\prod_{i=1}^t\alpha_t$ representing the noise schedules provided in \cite{ho2020denoising,song2020denoising}.

\begin{figure}[t]
\centering
\includegraphics[width=0.9\columnwidth]{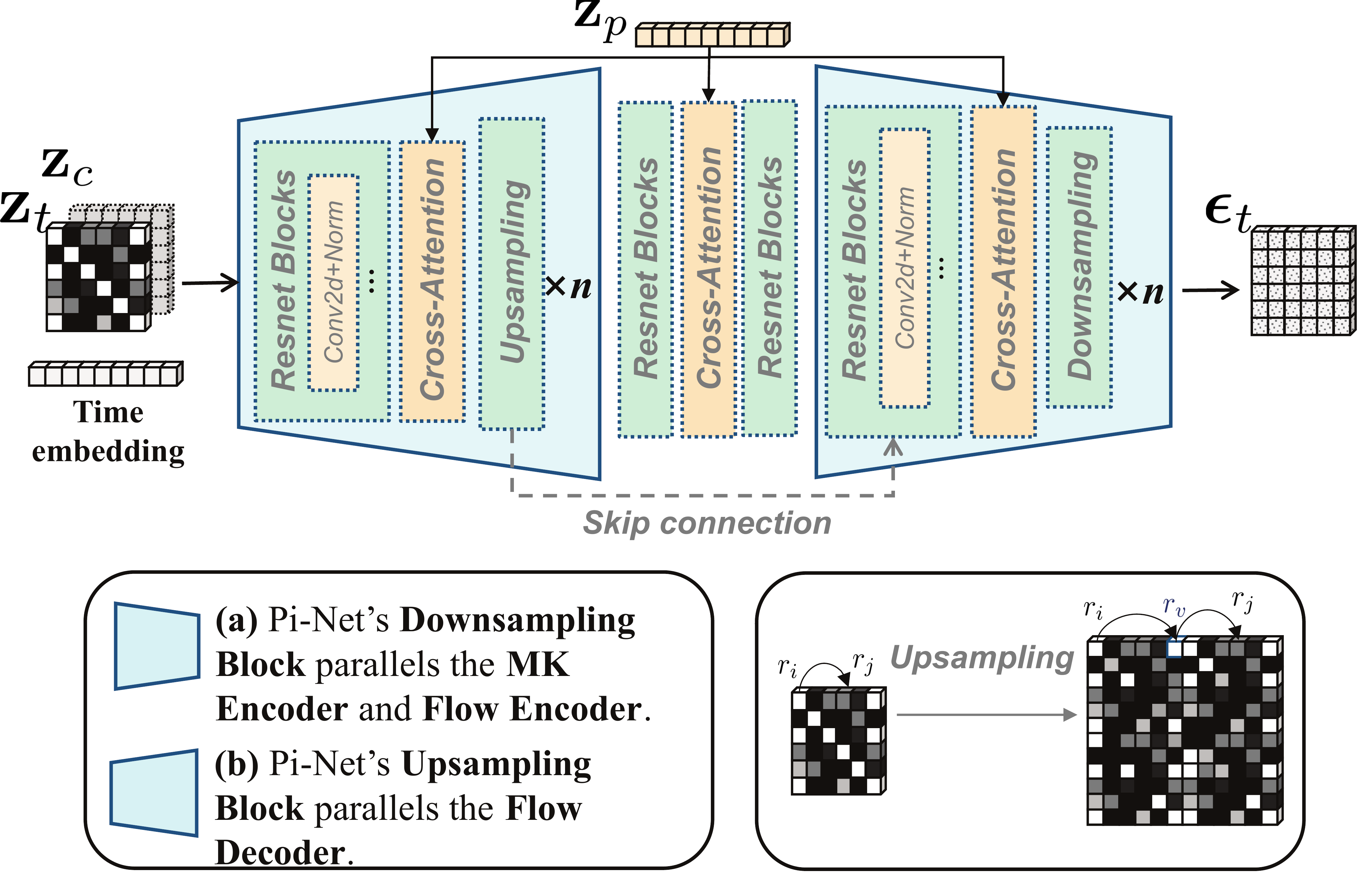}
\caption{Pi-Net Architecture: Inverted U-Net Design for OD Flow Generation.}
\label{fig:pinet}
\end{figure}

\textbf{Pi-Net}. We propose Pi-Net—a variant of U-Net with an inverted encoder-decoder structure that estimates noise $\boldsymbol{\epsilon}_\theta(\mathbf{z}_t, \mathcal{C}, t)$ through parameters $\theta$. 
Unlike conventional U-Net architectures that follow a downsample-then-upsample paradigm, Pi-Net adopts an up-then-down approach to address the unique characteristics of OD matrices, as illustrated in Figure~\ref{fig:pinet}. Specifically, the network first performs upsampling to expand spatial resolution, enabling explicit modeling of intermediate mobility patterns between regions (e.g., virtual transit hubs $r_v$ connecting $r_i$ to $r_j$), before applying downsampling to refine global coherence. 
This design ensures that latent codes at high resolutions retain fine-grained interaction details, which are critical for capturing spatial dependencies in OD flows. 

\textbf{Condition Injection}. The regional multi-kernel representation $\mathbf{z}_c$ and permutation embedding $\mathbf{z}_p$ are injected into the diffusion model through distinct mechanisms optimized for their respective characteristics. 
Given that $\mathbf{z}_c$ shares the same spatial resolution as the latent representation $\mathbf{z}_t$, it is concatenated along the channel dimension to provide spatially-aligned regional context. In contrast, $\mathbf{z}_p$ is integrated via a hierarchical cross-attention mechanism, acting as a structural prior to modulate spatial dependencies. 

\subsubsection{Equivariant Diffusion Training}

Our conditional latent diffusion module is trained with two complementary objectives. The primary objective enforces structural consistency across region index permutations: for each input sample, $N_p$ reindexed variants are generated via random permutations, and the model must reconstruct the corresponding OD matrices while properly aligning with the associated permutation embeddings. This is formalized as:
\begin{equation}
\mathcal{L}_\textrm{pre}=\frac{1}{N_p}\sum_{k=1}^{N_p}||\mathbf{M}_{\pi_k}-\mathcal{D}_{\pi_k}(\mathbf{z})||_2^2,
\end{equation}
where $\mathcal{D}_{\pi_k}(\mathbf{z})$ denotes the model's reconstructed output under permutation $\pi_k$.
The secondary objective maintains the standard diffusion loss for high-quality generation:
\begin{equation}
\mathcal{L}_\textrm{LDM}=\mathbb{E}_{\mathbf{z},\epsilon\sim\mathcal{N}(\mathbf{0},\mathbf{I}),t}\left[||\epsilon-\epsilon_\theta(\mathbf{z}_t,\mathcal{C},t)||_2^2\right].
\end{equation}

Following training, the Sat2Flow supports efficient sampling from the learned latent distribution and reconstructing OD flow matrices. The sampling process is accelerated using the DDIM method \cite{song2020denoising}, which reduces the number of denoising steps while maintaining generation quality, as detailed in Algorithm \ref{alg:generation}.

\begin{algorithm}[h]
    \caption{OD Matrix Generation through Sat2Flow}
    \label{alg:generation}
    \textbf{Input} ~~\\
    \qquad Regional satellite features $\{\mathbf{x}_i\}_{i=1}^N$ of a target city \\
    \qquad Permutation sequence $\pi(1),\pi(2),...,\pi(N)$ \\
    \qquad Length $\tau$ of subsequence in DDIM sampling\\
    \textbf{Output} ~~\\
    \qquad OD matrix $\mathbf{M}$ of that target city.
    \begin{algorithmic}[1]
    \STATE Compute multi-kernel condition $\mathbf{z}_c=\mathcal{E}(\mathcal{K})$ via multi-kernel encoder with $\{\mathbf{x}_i\}_{i=1}^N$
    \STATE Construct permutation embedding $\mathbf{z}_p$ based on $\pi$
    \STATE Sample $\mathbf{z}_T \sim \mathcal{N}(0, \mathbf{I})$ from the prior  
    \STATE Set step interval $\Delta t = \lfloor T / \tau \rfloor$  
    \FOR{$t = T, T-\Delta t, ..., \Delta t$}
        \STATE $\mathbf{z}_{t-\Delta t} \leftarrow \frac{1}{\sqrt{\alpha_t}} \left( \mathbf{z}_t - \frac{1-\alpha_t}{\sqrt{1-\bar{\alpha}_t}}\epsilon_\theta(\mathbf{z}_t, \mathcal{C}, t) \right)$
    \ENDFOR
    \STATE Reconstruct OD matrix via decoder: $\hat{\mathbf{M}}\leftarrow\mathcal{D}(\mathbf{z}_0)$
    \RETURN $\hat{\mathbf{M}}$
    \end{algorithmic}
\end{algorithm}

\section{Experiments}
\subsection{Datasets}
We evaluated our approach in the \textit{CommutingODGen} dataset~\cite{rong2024large}, which comprises 3,333 urban areas across the United States, including 3,233 counties and 100 metropolitan statistical areas. Each urban area is partitioned into fine-grained administrative units: census tracts for counties and census block groups for metropolitan regions. OD flow matrices are derived from the Longitudinal Employer-Household Dynamics Origin-Destination Employment Statistics (LODES), a comprehensive source maintained by the U.S. Census Bureau, capturing commuting patterns between these regional units.


\begin{table*}[t]
    \centering
    \scriptsize
        \begin{tabular}{@{\hspace{0.5pt}}c|ccc|ccc@{\hspace{0.5pt}}}
        \toprule
                & \multicolumn{3}{c|}{ Flow Value } & \multicolumn{3}{c}{ Empirical Distribution (JSD) }      \\ \midrule           
            \multicolumn{1}{c|}{Model}          & \multicolumn{1}{c}{CPC$\uparrow$} & \multicolumn{1}{c}{RMSE$\downarrow$} & \multicolumn{1}{c|}{NRMSE$\downarrow$} & \multicolumn{1}{c}{inflow$\downarrow$} & \multicolumn{1}{c}{outflow$\downarrow$} & \multicolumn{1}{c}{ODflow$\downarrow$}  \\ \midrule
        GM-P    & 0.321     & 174.0    & 2.222    & 0.668     & 0.656     & 0.409     \\
        GM-E    & 0.329     & 162.9    & 2.080    & 0.652     & 0.637     & 0.422     \\ \midrule
        RF      & 0.458     & 100.4    & 1.282    & 0.424     & 0.503     & 0.219     \\ 
        DGM     & 0.431     & 92.9     & 1.186     & 0.469     & 0.561     & 0.230     \\
        GMEL    & 0.440     & 94.3     & 1.204     & 0.445     & 0.355     & 0.207     \\
        NetGAN  & 0.487     & 89.1     & 1.138     & 0.429   & 0.354     & 0.191     \\
        DiffODGen & 0.532   & 74.6    & 0.953      & 0.324   & 0.270     & 0.149     \\ 
        WeDAN   & \underline{0.593} & \underline{68.6} & \underline{0.876} & \underline{0.291} & \underline{0.269} &  \underline{0.147}  \\ \midrule
        \multirow{2}{*}{\textbf{Sat2Flow}}   & \multicolumn{1}{c}{\multirow{2}{*}{\begin{tabular}[c]{@{}c@{}}\textbf{0.635}\\ \footnotesize{~(+7.06\%)}\end{tabular}}}    & \multicolumn{1}{c}{\multirow{2}{*}{\begin{tabular}[c]{@{}c@{}}\textbf{65.1}\\ \footnotesize{~(+5.14\%)}\end{tabular}}}     & \multicolumn{1}{c}{\multirow{2}{*}{\begin{tabular}[c]{@{}c@{}}\textbf{0.831}\\ \footnotesize{~(+5.14\%)}\end{tabular}}}     & \multicolumn{1}{c}{\multirow{2}{*}{\begin{tabular}[c]{@{}c@{}}\textbf{0.272}\\ \footnotesize{~(+6.52\%)}\end{tabular}}}   & \multicolumn{1}{c}{\multirow{2}{*}{\begin{tabular}[c]{@{}c@{}}\textbf{0.266}\\ \footnotesize{~(+1.10\%)}\end{tabular}}}     & \multicolumn{1}{c}{\multirow{2}{*}{\begin{tabular}[c]{@{}c@{}}\textbf{0.145}\\ \footnotesize{~(+1.36\%)}\end{tabular}}}     \\
        &           &          &           &           &           &          \\ \bottomrule
        \end{tabular}
    \caption{Performance comparison with baseline models on test cities.}
    \label{tab:performance}
\end{table*}

\begin{table}[t]
    \centering
    \scriptsize
        \begin{tabular}{@{\hspace{0.5pt}}c|ccccc@{\hspace{0.5pt}}}
        \toprule
            & \multicolumn{5}{c}{Permutation Increases (JSD $\downarrow$)}\\ \midrule
            \multicolumn{1}{c|}{Model} & \multicolumn{1}{c}{10\%} & \multicolumn{1}{c}{30\%} & \multicolumn{1}{c}{50\%} & \multicolumn{1}{c}{80\%} & \multicolumn{1}{c}{100\%} \\ \midrule
        GMEL    & 0.213     & 0.213     & 0.213     & 0.213     & 0.213\\
        NetGAN  & 0.206     & 0.210     & 0.221     & 0.225   & 0.231\\
        DiffODGen & 0.156   & 0.162     & 0.182     & 0.193   & 0.211\\ 
        WeDAN   & 0.158     & 0.167     & 0.179     & 0.193   & 0.207\\ \midrule
        \textbf{Sat2Flow}   & \textbf{0.149} &\textbf{0.154} & \textbf{0.160} & \textbf{0.164} & \textbf{0.171}\\ \bottomrule
        \end{tabular}
    \caption{Performance under progressive index permutation at varying intensities on test cities.}
    \label{tab:permutation}
\end{table}

\subsection{Baselines}
We evaluated the proposed Sat2Flow framework against seven representative baselines, categorized into two modeling paradigms: physics-based and data-driven models.
\textbf{Physics-based models} encompass the Gravity Model~\cite{zipf1946p} with power-law decay (GM-P) and exponential decay (GM-E), which rely solely on the population distribution as input and are grounded in the theoretical principles of human mobility.
\textbf{Data-driven models} include Random Forest (RF)~\cite{pourebrahim2018enhancing}, DeepGravity (DGM)~\cite{simini2021deep}, Geo-contextual Multitask Embedding Learning (GMEL)~\cite{liu2020learning}, NetGAN~\cite{bojchevski2018netgan}, and DiffODGen~\cite{rong2023complexity}, which leverage richer semantic and structural features. 


\subsection{Experimental Settings}
\subsubsection{Dataset Processing}
The dataset is partitioned into training, validation, and test sets using an 8:1:1 split, ensuring sufficient data for model learning and unbiased evaluation. A logarithmic transformation is applied to the OD flow matrices to address their heavy-tailed distribution. This step is motivated by theoretical studies that identify power-law scaling in human mobility patterns \cite{saberi2018complex}.

\subsubsection{Implementation Details}
The proposed Sat2Flow framework is implemented in PyTorch using the AdamW optimizer with a learning rate of $10^{-3}$ and weight decay of $10^{-2}$.
Contrastive learning and diffusion learning are trained for 1000 epochs with a batch size of 12. 
For data augmentation, we apply random permutations to regional indices at 50\% intensity to enhance robustness to index reordering. 
Satellite imagery is encoded into 768-dimensional features, while the MK encoder is configured with kernel dimension 64 and 8 kernels. 
All experiments were performed on a single NVIDIA GeForce RTX A6000 GPU with 48GB of memory.

\subsubsection{Evaluation Metrics}
To comprehensively assess the quality of generated OD matrices, we employ four complementary metrics that capture different aspects of generation fidelity.
\textbf{Flow-level accuracy} is measured using three error metrics that quantify deviations in flow values: Root Mean Square Error (RMSE), Normalized Root Mean Square Error (NRMSE), and Common Part of Commuting (CPC):
\begin{equation}
    \text{RMSE}=\sqrt{\frac{1}{N^2}\Vert\textbf{M}-\hat{\textbf{M}}\Vert_F^2},
\end{equation}
\begin{equation}
    \text{NRMSE}=\text{RMSE}\bigg/\sqrt{ \frac{1}{N^2} \Vert \mathbf{M} - \bar{\mathbf M}\Vert_F^2 },
\end{equation}
\begin{equation}
    \text{CPC}={ \frac{2\sum_{i,j}    \min([\mathbf{M}]_{i,j},[\hat{\mathbf{M}}]_{i,j})} {{\sum_{i,j}{[\mathbf{M}]_{i,j}}+ \sum_{i,j}{[\hat{\mathbf{M}}]_{i,j}}} }},
\end{equation}
where $\Vert\cdot\Vert_F$ denotes the Frobenius norm, and $\bar{\mathbf{M}}$ represents the mean of elements in the OD matrix $\mathbf{M}$.

\textbf{Distributional alignment} is quantified using Jensen-Shannon divergence (JSD) to measure global distributional similarity between real and generated OD matrices:
\begin{equation}
    \text{JSD}=\frac12D_{KL}(\mathbf{P}_{\mathbf{M}}\Vert\mathbf{P}_{\hat{\mathbf{M}}}) + \frac12D_{KL}(\mathbf{P}_{\hat{\mathbf{M}}}||\mathbf{P}_{\mathbf{M}}),
\end{equation}
where $\mathbf{P}_{\mathbf{M}}$ and $\mathbf{P}_{\hat{\mathbf{M}}}$ represent the marginal empirical distributions of the real matrix $\mathbf{M}$ and the generated matrix $\hat{\mathbf{M}}$, respectively.  

\subsection{Performance Comparison}
Table~\ref{tab:performance} presents a comprehensive performance comparison between Sat2Flow and seven representative baseline models across six evaluation metrics on test cities. The experimental results reveal the following key observations: 
(1) Sat2Flow achieves the highest CPC of 0.635 (+7.06\% over WeDAN) and lowest errors with RMSE of 65.1 and NRMSE of 0.831 (+5.14\% improvement). These substantial improvements demonstrate that, with our framework, satellite imagery alone captures underlying spatial mobility patterns more effectively than traditional feature-rich approaches.
(2) Our method achieves the lowest JSD across all mobility distributions.
The consistent improvements across local mobility characteristics and global connectivity patterns indicate that Sat2Flow effectively preserves the statistical properties of real mobility flows.
(3) Physics-based models show poor performance with CPC around 0.32, confirming their oversimplified assumptions, while data-driven models show progressive improvements. Notably, Sat2Flow surpasses even advanced feature-rich baselines, highlighting the efficacy of extracting mobility patterns directly from satellite imagery without region-specific auxiliary data.

\subsection{Consistency Evaluation}

To validate the structural consistency of our approach, we conduct an analysis of model performance under progressive index permutation at varying intensities (10\% to 100\%). Table \ref{tab:permutation} presents the distributional fidelity results measured through JSD. It is evident that pair-wise level models, represented by GMEL, inherently adapt to different index orderings as they predict flows for individual regions in isolation, with consistent JSD values of 0.213 across all permutation intensities; however, their accuracy remains limited due to the neglect of spatial interaction information. In contrast, recent matrix-wise level models, which generate city-scale multi-regional flows in a single step, capture spatial structures but exhibit sensitivity to isomorphic spatial configurations (i.e., index permutations), with performance deteriorating as permutation intensity increases. Sat2Flow demonstrates minimal performance variation under index permutations while maintaining high accuracy, attributable to its permutation-equivariant diffusion training.

\begin{figure}[t]
\centering
\includegraphics[width=\columnwidth]{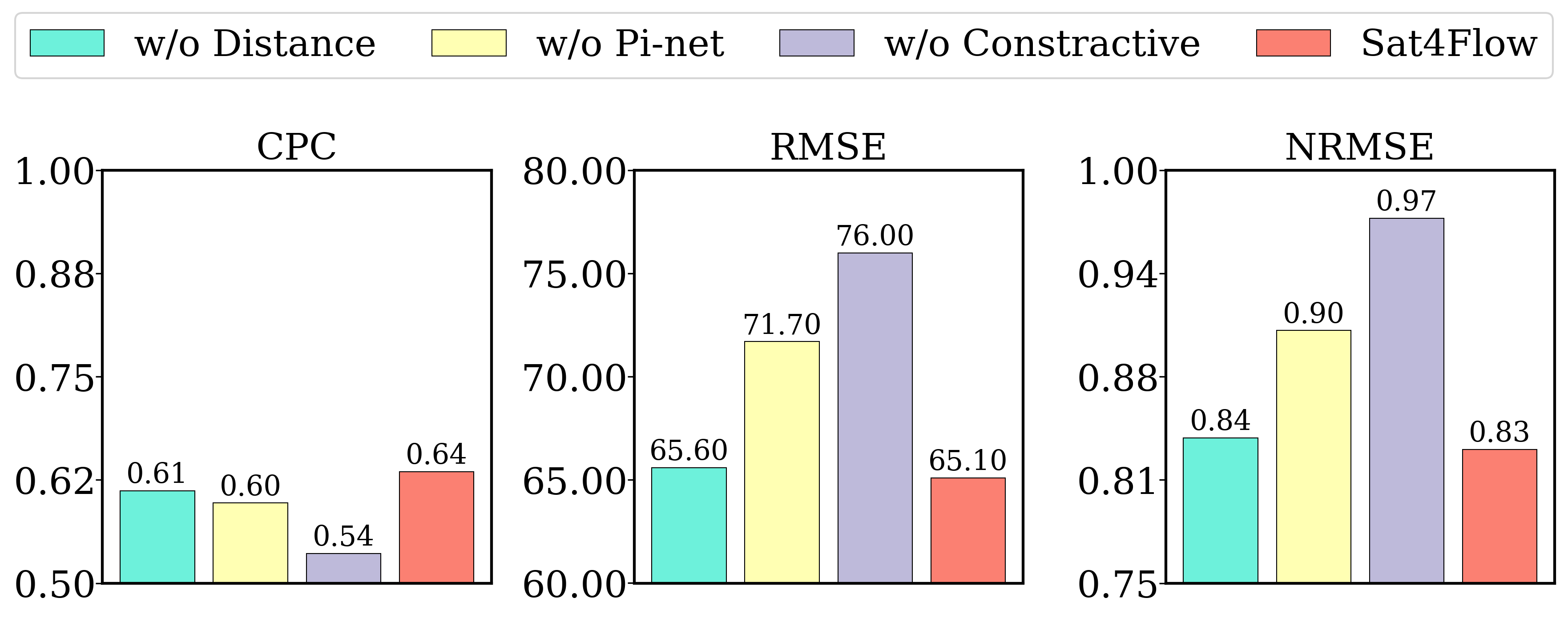}
\caption{Ablation study evaluating component contributions through CPC, RMSE, and NRMSE metrics.}
\label{fig:ablation}
\end{figure}

\begin{figure}[t]
\centering
\includegraphics[width=\columnwidth]{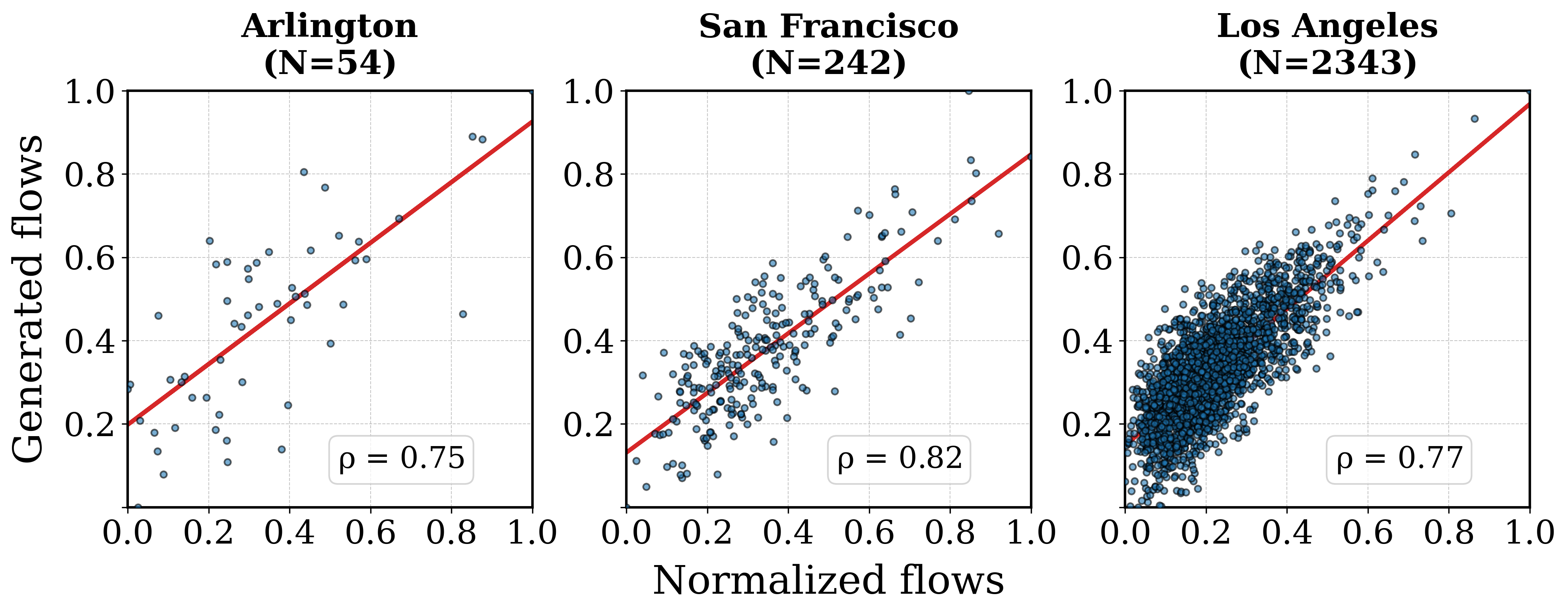}
\caption{Case study analyzing model behavior across selected cities with diverse sizes.}
\label{fig:case}
\end{figure}

\subsection{Ablation Study}
To systematically evaluate the contribution of each component in Sat2Flow, we design three ablation variants by removing specific modules: \textbf{\textit{w/o} Distance}: Eliminates the topological distance prior in multi-kernel encoding, removing explicit spatial relationship constraints. \textbf{\textit{w/o} Pi-net}: Replaces the permutation-aware Pi-net backbone with a standard U-net architecture. \textbf{\textit{w/o} Contrastive}: Removes the contrastive learning phase, disabling latent space alignment between regional representations and OD flows during the first training stage.
As shown in Figure~\ref{fig:ablation}, the full Sat2Flow consistently outperforms all ablated variants across all metrics. 
\textit{w/o} Distance achieves 65.60 RMSE, only 0.8\% higher than full Sat2Flow, confirming that the multi-kernel encoder effectively captures regional relationships while demonstrating that satellite imagery inherently encodes spatial relationships without explicit distance modeling. 
\textit{w/o} Pi-net shows significant degradation, demonstrating that Pi-net's up-then-down architecture is essential for preserving intermediate mobility patterns that standard U-net misses. 
\textit{w/o} Contrastive suffers the largest performance drop as eliminating the contrastive learning phase prevents the condition embeddings from being guided into a latent space aligned with OD semantic representation, forcing the model to search within a broader solution space from scratch and consequently converge to suboptimal local minima.

\subsection{Case Study}
We evaluate model performance across cities of varying scales by comparing generated and real OD flows, as shown in Figure~\ref{fig:case}. 
Three representative cities were selected: Arlington, San Francisco, and Los Angeles. It can be observed that model performance peaks at medium-sized cities ($\rho=0.82$ for San Francisco), while slightly declining for both smaller ($\rho=0.75$ for Arlington) and larger cities ($\rho=0.77$ for Los Angeles).
This pattern aligns with the concept of urban scaling laws~\cite{pumain2006evolutionary}, where medium-sized cities exhibit balanced spatial heterogeneity—not constrained by limited connectivity in small towns nor overwhelmed by excessive functional fragmentation in megacities. The consistent performance across scales confirms Sat2Flow's robustness to urban morphology variations, while the minor dip in megacities reflects challenges in modeling long-tail mobility patterns across highly diverse areas.

\section{Related Work}

Human flow generation has progressed through distinct methodological phases, transitioning from labor-intensive data collection to automated modeling approaches~\cite{luca2021survey}. 
Early approaches relied on manual travel surveys and census data collection~\cite{axhausen2002observing, iqbal2014development}, which were prone to sampling bias and substantial respondent burden. 
Subsequent physical modeling approaches, including the gravity model~\cite{zipf1946p} and the radiation model~\cite{simini2012universal}, model human mobility through population-based interactions. However, their exclusive reliance on demographic counts oversimplifies urban complexity and fails to capture nuanced socioeconomic mobility patterns inherent in modern cities~\cite{cao2025disentangling}.
The emergence of big data enabled more sophisticated modeling through comprehensive urban descriptors. Tree-based methods~\cite{pourebrahim2019trip} achieved initial success, while deep learning architectures demonstrated initial success by incorporating sociodemographics, POIs, and land use patterns. Deep learning architectures further enhanced prediction accuracy by leveraging road network topology and multi-modal urban features~\cite{wang2025st}.
Recent advances have focused on explicitly modeling spatial relationships through graph neural architectures~\cite{pourebrahim2018enhancing, liu2020learning, yao2020spatial}.
Despite these advances, deployment remains challenging in data-scarce regions~\cite{haraguchi2022human, cao2025urban}. More critically, no existing methodology maintains structural consistency under regional index permutations—a fundamental requirement for robust city-scale mobility synthesis that Sat2Flow addresses through its topology-agnostic design.

\section{Conclusion}
This work introduces Sat2Flow, a novel framework that generates high-fidelity OD flow matrices using only satellite imagery without requiring auxiliary urban data. By integrating multi-kernel encoding with contrastive learning and permutation-aware diffusion modeling, our approach preserves structural coherence under arbitrary regional index permutations while capturing both regional characteristics and global mobility patterns. Extensive experiments demonstrate that Sat2Flow outperforms state-of-the-art methods across multiple evaluation metrics, achieving 7.06\% higher CPC and 6.52\% lower JSD under index permutations compared to the strongest baseline. The framework's ability to leverage universally accessible remote sensing data while maintaining structural consistency offers a scalable solution for urban mobility analysis in data-scarce regions.

\section{Acknowledgments}
This research was supported by Shenzhen Science and Technology Program (No. JCYJ20240813113300001, 20231127180406001).

\bibliography{aaai2026}
\end{document}